\documentclass[cameraready]{Interspeech}
\usepackage{soulutf8}

\title{Synchronization and Turn-Taking in Full-Duplex Speech Dialogue Models}

\author[affiliation={1,2, \dagger}, orcid=0000-0000-0000-0000, equalcontribution]{Pablo}{Riera}
\author[affiliation={1, \dagger}, orcid=0000-0000-0000-1111, equalcontribution]{Pablo}{Brusco}
\author[affiliation={1}, orcid=0000-0000-0000-2222]{Cristina}{Kuo}
\author[affiliation={1}, orcid=0000-0000-0000-3333]{Marcelo}{Sancinetti}
\author[affiliation={1}, orcid=0000-0000-0000-4444, correspondingauthor]{S.R.K.}{Branavan}

\address{
    $^1$ ASAPP Inc., USA \\
    $^2$ Departamento de Computación, FCEyN, Universidad de Buenos Aires, Argentina
}

\email{priera@dc.uba.ar, \{pbrusco, ckuo, msancinetti, branavan\}@asapp.com\\
$\dagger$ This work was performed while the author was at ASAPP.}

\keywords{full-duplex spoken dialogue, Speech LLMs, entrainment, synchronization, turn-taking, neural coupling,  simulated dialogue}

\usepackage{comment}

\begin{document}

\maketitle

\begin{abstract}
Full-duplex spoken dialogue models (SDMs) can listen and speak simultaneously, enabling interaction dynamics closer to human conversation than turn-based systems. Inspired by neural coupling in human communication, we study how such models coordinate their internal representations during interaction.
We simulate full-duplex dialogues between two instances of the pretrained \textit{Moshi} model under controlled conditions, manipulating channel noise and decoding bias. Synchronization is measured using Centered Kernel Alignment (CKA) across temporal lags, while anticipatory turn-taking cues are probed from delayed internal activations using causal LSTM models, from both speaker and listener perspectives.
We find strong representational synchronization under no noise conditions, peaking near zero lag and degrading with noise, and we show that internal states encode anticipatory information that supports turn-taking prediction ahead of time.
\end{abstract}

\section{Introduction}

Human conversation is characterized by rich temporal dynamics in which participants coordinate timing, mirror prosody, and anticipate turn changes \cite{duncan1974structure, stivers2009universals}. These phenomena include what is referred to as \textit{entrainment}, the tendency of interlocutors to converge in speech rate, pitch, and syntactic structure, although they may also adapt by becoming more dissimilar while remaining responsive, a process known as \textit{disentrainment} \cite{galvez2020empirical}. In spoken dialogue systems (SDS), perceived interaction naturalness is closely linked to the extent to which such human-like patterns can be reproduced \cite{stivers2009universals, skantze2021turn}.

Beyond observable speech features, successful communication is supported by \textit{neural coupling} between speaker and listener. fMRI studies show that their brain activity becomes temporally and spatially aligned, and that this alignment disappears when communication fails or the language is unintelligible. Moreover, stronger \textit{anticipatory} coupling—where listener activity precedes the speaker’s—is associated with improved comprehension, highlighting prediction as a core mechanism of communication \cite{stephens2010coupling}.

Recent advances in Full-Duplex Spoken Dialogue Systems (FDSDS), which can listen while speaking, have renewed interest in modeling and evaluating turn-taking behavior. Inspired by findings on neural coupling in human communication, we propose an assessment framework that studies synchronization and turn-taking in full-duplex dialogue models. We instantiate this approach using \textit{Moshi}, a pretrained \textit{full-duplex} speech dialogue model, by simulating dialogues between two model instances and analyzing their internal dynamics. Synchronization is quantified through alignment of internal representations across time lags \cite{stephens2010coupling,wilmer2012time}, while turn-taking information is examined by probing internal states with LSTM-based models to predict turn boundaries and transitions \cite{brusco2023automatic}.

\section{Related Work}

Spoken Dialogue Systems (SDS) have shifted from rigid, turn-based half-duplex architectures toward synchronous, full-duplex interactions that more closely resemble human communication. Traditional cascaded pipelines for speech recognition, language modeling, and synthesis suffer from latency and unnatural turn-taking \cite{Yang2025TowardsHE, Peng2025FDBenchAF}. To address these issues, recent work has focused on Full-Duplex SDS (FDSDS), including streaming end-to-end models such as Moshi and Mini-Omni, as well as voice-activity-augmented systems like Freeze-Omni and VITA \cite{Chen2025FromTT, Ge2025FLEXIBF}. A key milestone was the introduction of Generative Spoken Dialogue Language Modeling (dGSLM), which showed that turn-taking, backchanneling, and overlap can be learned directly from raw audio without textual supervision \cite{nguyen2023generative, Wang2025TowardsGA}.

Among end-to-end approaches, \emph{Moshi} \cite{defossez2024moshispeechtextfoundationmodel} is a pretrained speech-text foundation model designed for real-time, bidirectional dialogue. Its architecture supports simultaneous conditioning on user input and self-generated speech, enabling interruptions and backchannels without explicit turn segmentation. While prior work has examined Moshi’s observable behaviors, its internal representation dynamics during interaction remain largely unexplored. 

As full-duplex models mature, benchmarking efforts have increasingly focused on behavioral evaluation. Benchmarks such as Full-Duplex-Bench \cite{lin2025full} and Talking Turns \cite{arora2025talking} assess conversational timing and backchannel appropriateness, with \cite{arora2025talking} arguing that corpus-level statistics fail to capture fine-grained interactional dynamics and proposing judge models trained on human data instead.

These benchmarks primarily rely on human recordings and largely neglect entrainment, synchronization, and internal turn-taking dynamics. While one option is to analyze human–system interactions, we instead connect full-duplex systems directly through a communication channel, enabling controlled yet naturalistic observation of emerging coordination. This approach not only advances the analysis of speech dialogue models, but also positions them as computational proxies for studying the principles underlying human turn-taking and coordination.

\section{Methods}
\label{sec:methods}

\subsection{Simulated Full-Duplex Dialogues Environment}
\label{sec:dialogues}

To study synchronization between internal representations and turn-taking dynamics under controlled conditions, we generate conversations between two independent instances of Moshi of 100 seconds. As the model can generate speech while consuming audio we connected two instances (A and B) through a communication channel based on direct token-level audio routing: the output audio tokens of Model A are fed as input to Model B, and vice versa. We manipulated this channel by including random token changes. For each frame we use a probability value to allow tokens to be altered and at values of 0.7 probability the speech becomes unintelligible. 

Conversations are initiated with a prerecorded audio prompt to induce the models to adopt the roles of either an agent or a customer in a medical appointment request scenario. We use the official Moshi checkpoints: ``Moshika'' for the agent role and ``Moshiko'' for the client role. To study the effect of task adaptation, we compare the default checkpoint with a fine-tuned version, allowing us to analyze the impact of pretraining versus fine-tuning on synchronization and turn-taking dynamics.

We further manipulate turn-taking behavior by introducing a decoding bias on the PAD token, which is emitted when the model is silent. Specifically, we subtract a constant value from the PAD token logits, increasing the likelihood of speech and thereby promoting more frequent turn-taking.

Experimental conditions include four noise levels (none, low, medium, high), two model versions (default and fine-tuned), yielding four agent–client pairings (default/default, default/fine-tuned, fine-tuned/default, fine-tuned/fine-tuned), and three PAD bias levels (none, medium, high). For each condition, we generate twenty dialogues with different random seeds, resulting in 2880 conversations totaling approximately 80 hours of audio.

To analyze internal dynamics and interaction timing, we extract activations from the final layer of the temporal transformer at each frame, denoted as $\mathbf{h}_t \in \mathbb{R}^d$. These activations are highly contextualized and directly used to generate speech and text at the next time step. For temporal annotation, we compute the onset and offset of interpausal units (IPUs), defined as voiced segments separated by at least 80 ms of silence. IPU boundaries define the decision points for the end-of-IPU (EOI) and Hold vs.\ Non-Hold tasks described below.

\subsection{Internal Representation Synchronization}

Synchronization between the two models is quantified using linear CKA computed between their internal representation time series at different temporal lags. CKA is a similarity measure invariant to orthogonal transformations and isotropic scaling \cite{kornblith2019similarity}. Linear CKA is defined as
\[
CKA(X,Y) =\frac{\lVert Y^\top X \rVert_F^2}{\lVert X^\top X \rVert_F \, \lVert Y^\top Y \rVert_F},
\]
where $X$ and $Y \in \mathbb{R}^{d \times n}$ are matrices of activations over $n$ frames with dimensionality $d$, and $\lVert \cdot \rVert_F$ denotes the Frobenius norm. Higher values indicate greater similarity between representation spaces.

In typical dialogue, only one speaker is active at a time; thus, high CKA values indicate alignment between models despite differing roles (speaking vs.\ listening). By computing CKA across temporal lags, we assess the degree of synchronization between the two models \cite{wilmer2012time}. We expect CKA to peak near zero lag for synchronized interactions and to decrease as lag increases. Increased noise in the communication channel is expected to reduce CKA values. We also computed mutual information between representation time series, but observed similar trends and therefore omit these results.

\subsection{Turn-taking Probing Setup}
\label{sec:methods:probing}

To test whether turn-taking information is encoded in the model’s internal representations, we employ a probing framework. We extract internal activations $\mathbf{h}_t \in \mathbb{R}^d$ from the final layer of the temporal transformer at each time frame $t$ and use them as input features for predictive probes trained to forecast upcoming turn-taking events. To enforce causality, we introduce a temporal delay $\delta$ such that the probe at time $t$ observes only the past representation $\mathbf{h}_{t-\delta}$.

Turn-taking events occur at specific time points $b$, corresponding to IPU boundaries. For each task, we train separate probes using delayed representations. We consider two perspectives: (i) speaker-side (production), where probes use the model’s own delayed states to predict its upcoming turn transitions, and (ii) listener-side (perception), where probes use the model’s own delayed states to predict the partner’s upcoming transitions. This distinction allows us to examine whether turn-taking cues are present in both planning and perceptual representations.

We expect the production setting to be easier, as the model has direct access to its own planning states. The perception setting is more challenging, as it requires inferring turn-taking intentions from externally observable cues. Nevertheless, if internal representations are aligned, we expect listener-side probes to perform above chance.

We consider two tasks: (i) end-of-IPU prediction and (ii) Hold vs.\ Non-Hold classification. For both tasks, probes trained on shuffled labels are used to estimate chance performance.

\subsubsection{End-of-IPU Prediction}

This task evaluates whether upcoming IPU-final boundaries are encoded in internal representations. It is formulated as a continuous frame-level prediction problem, where the probe produces an output at every time frame.

Positive labels are assigned only to true EOI frames (the final voiced frame before at least 80 ms of silence), while all other frames are labeled as negative. The classifier must therefore distinguish genuine turn-final offsets from speech, micro-pauses, and within-IPU fluctuations based on delayed representations. An example is shown in Figure~\ref{fig:targets}.

Formally, we use a recurrent probe that at each time index $t$, observes the signal up to a time offset $\delta$ before the target, i.e., the hidden state $\mathbf{h}_{t-\delta}$ and outputs a probability $y_t$ that the frame corresponds to an EOI.

This experiment tests whether internal representations encode anticipatory information about imminent IPU boundaries, independent of whether the speaker ultimately continues (Hold) or yields the floor, and how early such cues become decodable.

\subsubsection{Hold vs. Non-Hold Prediction}

This task targets higher-level conversational decisions. At IPU boundaries, the speaker either continues speaking (Hold) or a turn shift occurs (Non-Hold), where the interlocutor takes the floor (e.g., smooth switches, interruptions, or butting-ins).

\begin{figure}
    \centering
    \includegraphics[width=1\linewidth]{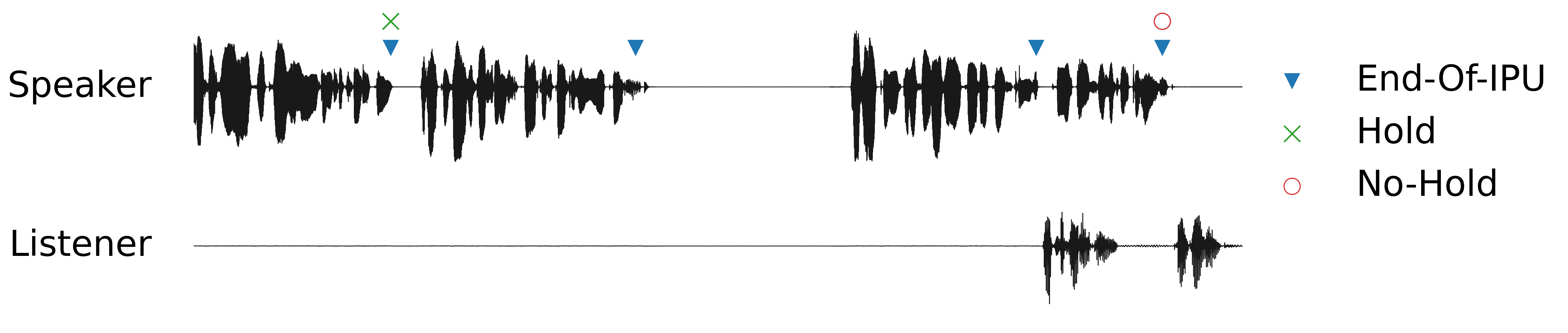}    
    \caption{Targets locations for the two turn-taking prediction tasks. End-of-IPU prediction is a continuous frame-level task, with positive labels only at true IPU-final frames (triangles). Hold vs.~Non-Hold prediction occurs at discrete IPU boundaries, classifying whether the speaker will continue (Hold as cross) or transition (Non-Hold as circle).}
    \label{fig:targets}
\end{figure}

Transitions are excluded if the pause to the next IPU exceeds one second or if no clear Hold/Non-Hold decision can be made, defined as both speakers overlapping for more than 240 ms. Shorter overlaps are labeled as Non-Hold, reflecting the common overlap observed in natural human and full-duplex interactions. Larger overlaps are excluded. Figure~\ref{fig:targets} illustrates these cases, including EOIs followed by long pauses, and EOIs with excessive overlap.

The probe is fed similarly to the EOI case, but the predictions are only made at valid turn transitions. We hypothesize that, on the speaker side, Hold vs.\ Non-Hold decisions involve planning across multiple IPUs and are reflected in the latent trajectory preceding the boundary. On the listener side, turn-taking cues may be inferred from prosodic or content-based signals in the partner’s speech, indicating an imminent turn transfer.

\subsubsection{Training Details}

We implement the probe as a causal LSTM with hidden size $H=64$, followed by a linear projection applied at every time step. Given a sequence $(\mathbf{h}_{1}, \dots, \mathbf{h}_{T-\delta})$, the probe produces per-frame $\mathbf{y}_t$ logits

$$
\mathbf{y}_t = \mathbf{W} \cdot \mathrm{LSTM}(\mathbf{h}_{t-\delta}) + \mathbf{b}
$$

Binary cross entropy loss is used. For the End-of-IPU task, loss is computed at every frame; for the Hold vs.\ Non-Hold task \cite{brusco2023automatic}, loss is computed only at valid targets frames. We train for a fixed number of epochs (200) using Adam optimizer with learning rate $10^{-3}$ and batch size 16.

This setup ensures strict causality: for any potential boundary, the model has access only to past internal states. We report AUC-ROC as the primary metric, evaluated as a function of delay $\delta$ to quantify the temporal reach.

Each dialogue is considered a data instance. To test our hypothesis based on channel noise, finetuning and bias conditions (Section \ref{sec:dialogues}) we train a probe for each combinations giving a total of 40 dialogues per probe because a model configuration will engage in dialogues with two different interlocutors (default and finetuned) and twenty random seeds. We split 32 for train and 8 for testing. No hyperparameter tuning was performed. To be sure the probe is actually learning, we also train probes using shuffled labels to estimate chance performance.

\section{Experiment Results}
We design two sets of experiments to investigate synchronization and turn-taking dynamics. All experiments are conducted in the simulated two-agent environment described in Section~\ref{sec:methods}. First we evaluate the emergence of synchronization under different communication conditions: (i) varying communication channel noise levels, (ii) changing pad token bias, and (iii) comparing pretrained vs.\ fine-tuned models. Second, we probe the internal representation to assess (i) whether upcoming IPU boundaries can be predicted from delayed activations, (ii) whether higher-level turn-management decisions such as hold vs. non-hold are encoded. For all predictive tasks, we evaluate both the production-side (speaker) and perception-side (listener) settings, and analyze how far in advance such information can be decoded from delayed internal representations.

\subsection{Synchronization of Internal Representations}
\label{sec:exp:synch}

\begin{figure}
    \centering
    \includegraphics[width=1\linewidth]{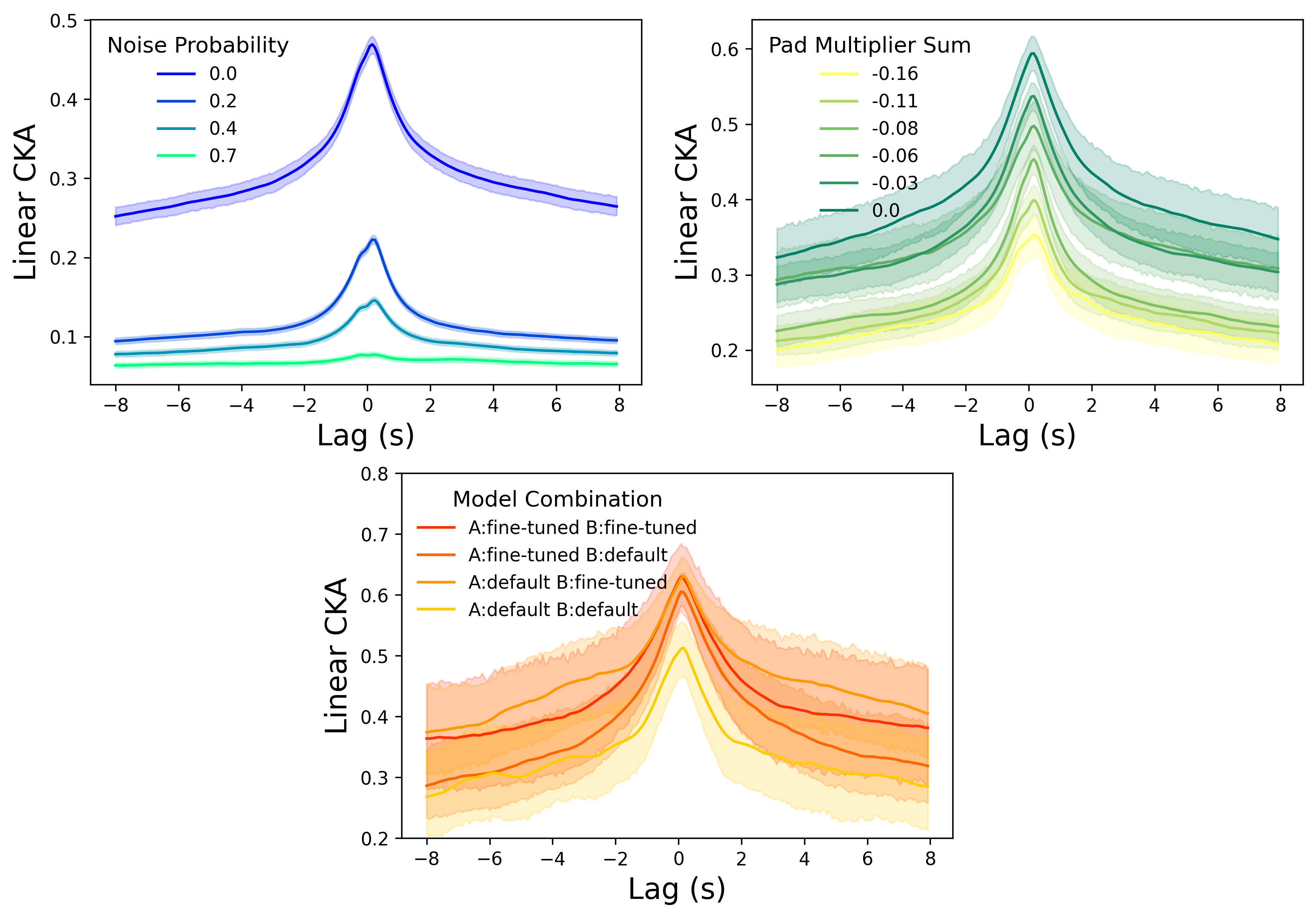}
    \caption{Linear CKA between the internal representations of the two agents for different lags and communication channel noise. Higher CKA values indicate stronger synchronization. Shaded regions represent 95\% confidence intervals of the mean CKA value.}
    \label{fig:synch}
\end{figure}

Results are shown as curves of linear CKA between the internal representations of the two agents at different temporal lags. Negative lags correspond to agent B leading agent A, while positive lags correspond to agent A leading agent B. In top left panel of Figure \ref{fig:synch} we see the curves for different noise levels in the communication channel. The main result is a sharp peak in the range of -2 to 2 seconds, indicating strong synchronization. The peak CKA magnitude is around 0.5 on average, but can be as large as 0.8 for some particular conversations. Notably, the baseline CKA value at large delays and under zero noise condition (value approx 0.25) is higher than under high-noise conditions (values less than 0.1), indicating that there is some internal representation alignment between the two models even when they are artificially delayed. This likely reflects the fact that the models still engage in a coherent conversational exchange with shared pacing, which is not the case under high-noise conditions, where each model effectively speaks in isolation while receiving mostly noise from the other side.


In the top right panel of Figure~\ref{fig:synch}, we show the curves for different PAD token biases under zero-noise conditions. As we paired models A and B that can have different bias values, we show the result of summing both values in the legend. More negative values indicate that the PAD token is less likely to be emitted, causing the models to speak earlier than in the unbiased case. We see the CKA values decrease with the bias, indicating that model alignment decreases when the decoding process is artificially altered and the conversation becomes less predictable.


Finally, in bottom panel of Figure \ref{fig:synch} we see the curves for different model types and conversations pairs. The main observation is that the combination "A:Default B:Default" has a lower peak value and the combinations that include the fine-tuned version have a higher peak value. This may be because the fine-tuned model was exposed during training to conversations similar in topic to those used in the experiments. 

\subsection{End-of-IPU Prediction}

\begin{figure}
    \centering
    \includegraphics[width=\linewidth, trim=0 0 0 8, clip]{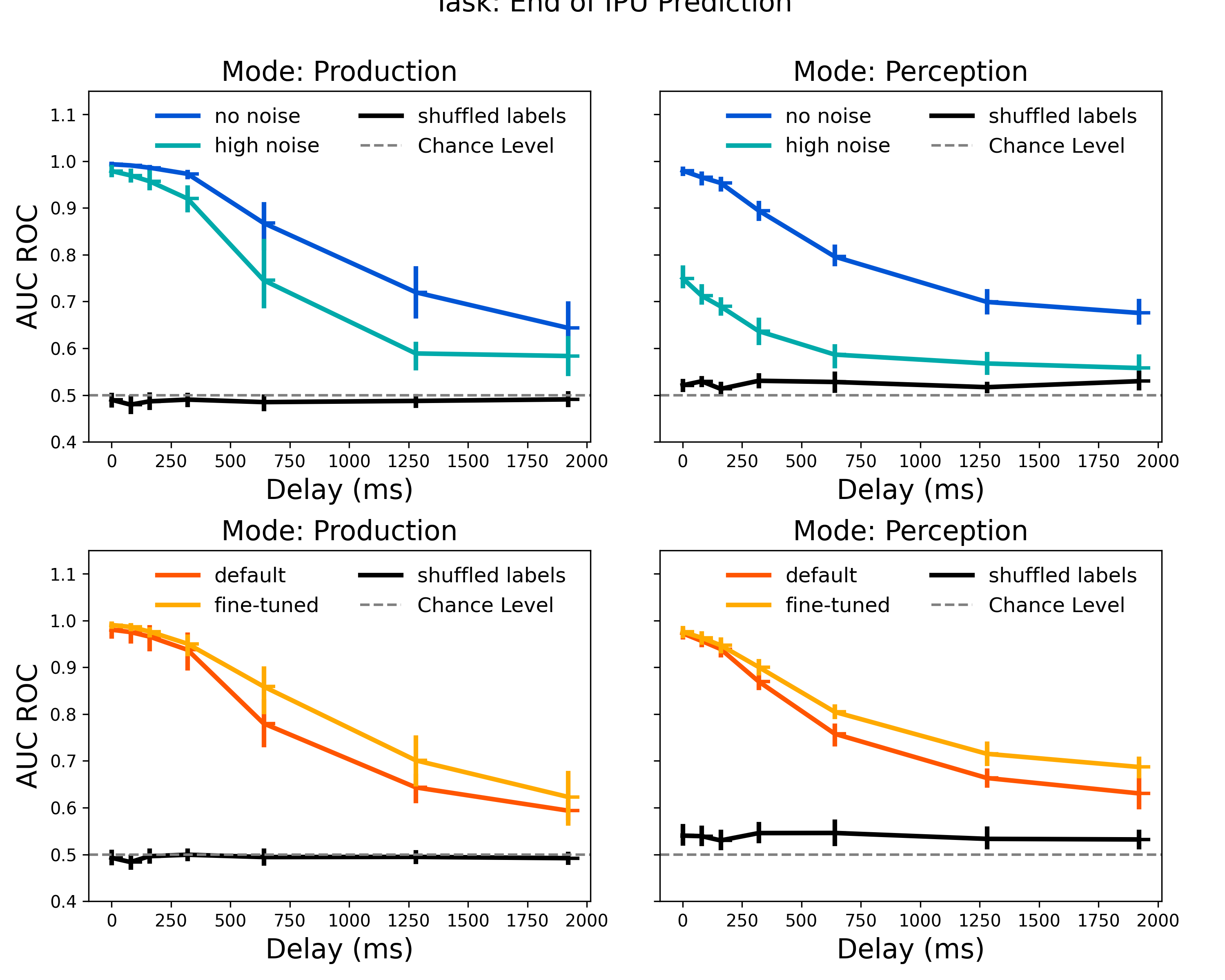}
    \caption{End-of-IPU prediction performance (AUC-ROC) across temporal delays and experimental conditions. Left and right panels correspond to production and perception settings, respectively. Top and bottom panels show noise vs.\ no-noise conditions and model type. Error bars represent 95\% confidence intervals.}
    \label{fig:endofipu}
\end{figure}

This experiment evaluates whether internal representations encode information about imminent IPU boundaries. Using the probing setup described in Section~\ref{sec:methods:probing}, we analyze predictive performance as a function of temporal delay ($\delta \in \{0,\ldots,1920\}$ ms) for both production (speaker-side) and perception (listener-side) representations, comparing noise vs.\ no-noise conditions and default vs.\ fine-tuned models.

The top two panels of Figure~\ref{fig:endofipu} show that, in both production and perception settings, the no-noise condition yields consistently higher AUC-ROC values across all delays than the noisy channel, indicating that clear communication strengthens the encoding of turn-final cues in internal representations.

The bottom two panels show a smaller performance gap between default and fine-tuned models, suggesting that fine-tuning has only a minor effect on EOI predictability.

\subsection{Hold vs.\ Non-Hold Turn-Management Prediction}

\begin{figure}
    \centering
    \includegraphics[width=1\linewidth, trim=0 0 0 8, clip]{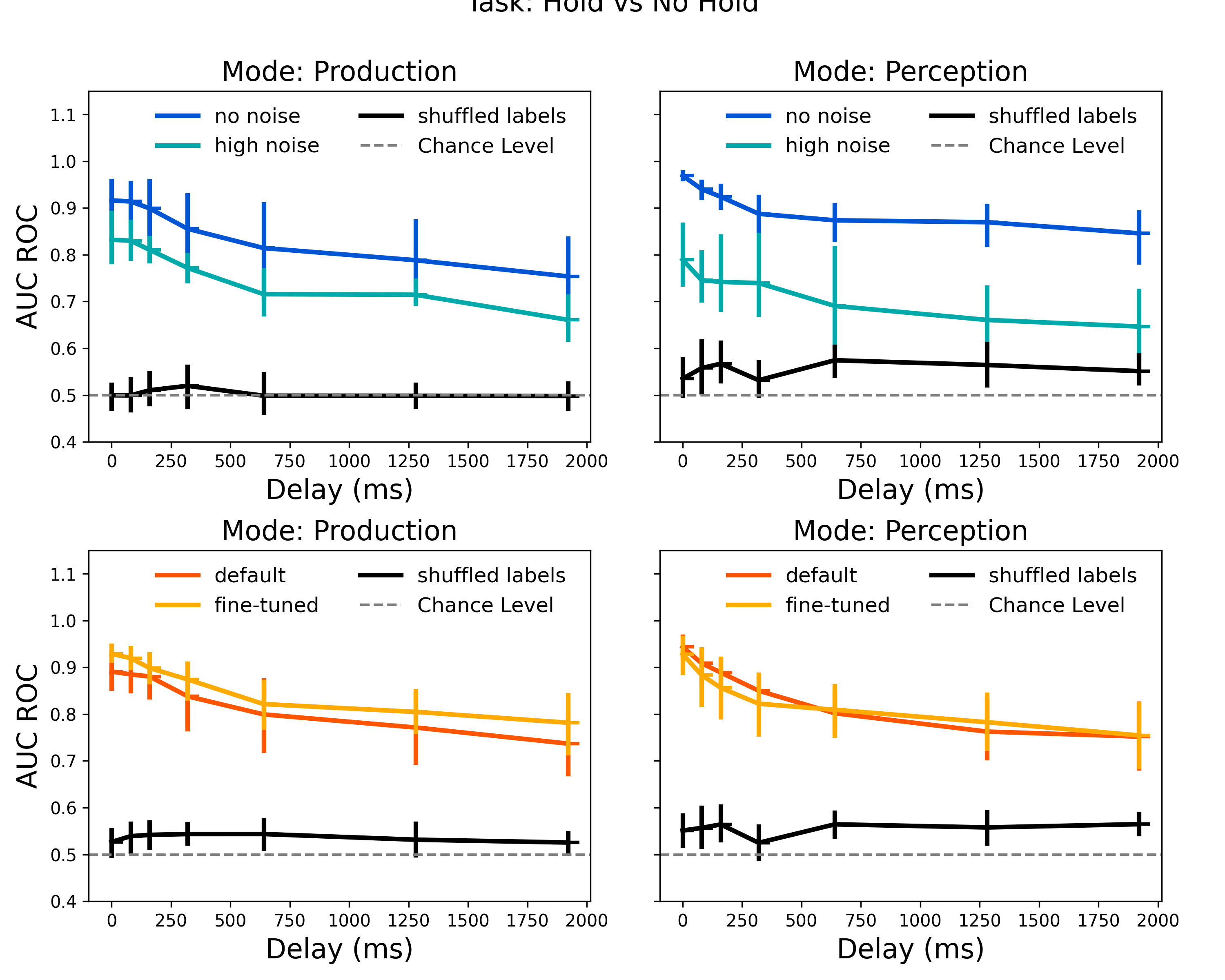}
    \caption{Hold vs Non-Hold prediction performance (AUC-ROC) across temporal delays and experimental conditions. Left and right panels correspond to production and perception settings, respectively. Top and bottom panels show noise vs.\ no-noise conditions and model type. Error bars represent 95\% confidence intervals.}
    \label{fig:holdvsnonhold}
\end{figure}

This experiment tests whether higher-level turn-management decisions (Hold vs.\ Non-Hold) are encoded in the model’s internal representations. We evaluate this task across conditions using the probing framework described in Section~\ref{sec:methods:probing}.

As shown in the top two panels of Figure~\ref{fig:holdvsnonhold}, the no-noise condition again leads to higher AUC-ROC values across delays in both production and perception settings, indicating stronger encoding of turn-management cues under clearer communication conditions. Compared to the End-of-IPU task, performance degrades more slowly with increasing delay. The bottom two panels show negligible differences between default and fine-tuned models, consistent with the EOI results.

Due to space limitations, turn-taking prediction results for PAD token bias are not shown. We observed AUC-ROC drops for EOI prediction in the production setting and for Hold vs.\ Non-Hold prediction in the perception setting.

Across both tasks, performance does not drop to chance level even at longer delays. We interpret this as an optimistic estimate, likely due to the limited diversity and shared temporal structure of the simulated dialogues, which are based on the same prompt and relatively short interactions. Under these conditions, probes may exploit global temporal regularities rather than relying exclusively on informative internal representations at larger delays. We expect this effect to diminish in more diverse conversational settings, which we leave for future work.

\section{Discussion and Future Work}

Our findings demonstrate that internal representational synchronization in full-duplex models emerges as a computational analog to human neural coupling \cite{stephens2010coupling}, showing high sensitivity to communication channel integrity. The alignment captured via CKA, combined with the probes' ability to forecast turn transitions with significant temporal lead, confirms that these models encode robust anticipatory cues for dialogue management from both production and perception perspectives. This framework moves beyond surface-level metrics to quantify the ``interactional health'' of end-to-end speech systems, positioning them as effective proxies for exploring the fundamental computational principles of human coordination and turn-taking behavior.

Our study is primarily limited by the narrow scope of conversational scenarios, which may introduce task-specific biases in probing performance. Future research should explore combinations of other full-duplex models, and layer-wise dynamics to pinpoint where synchronization emerges within the architecture and validate these findings in human--model interactions to ensure that representational coupling generalizes beyond simulated environments.





\bibliographystyle{IEEEtran}
\bibliography{mybib}

\end{document}